\documentclass[conference]{IEEEtran}
\usepackage[dvips]{graphicx}
\usepackage{subfigure}
\usepackage{color}

\ifCLASSINFOpdf
\else
\fi
\begin{document}
%
\title{Semantic Change Detection with Hypermaps}


\author{\IEEEauthorblockN{Teppei Suzuki (AIST,Keio), Soma Shirakabe (AIST, Tsukuba Univ.), Yudai Miyashita (Tokyo Denki Univ.)}
\IEEEauthorblockN{Akio Nakamura (Tokyo Denki Univ.), Yutaka Satoh (AIST), Hirokatsu Kataoka (AIST)}
\IEEEauthorblockN{Email: hirokatsu.kataoka@aist.go.jp\\
http://hirokatsukataoka.net/}
}


%


\maketitle

\begin{abstract}
Change detection is the study of detecting changes between two different images of a scene taken at different times. By the detected change areas, however, a human cannot understand how different the two images. Therefore, a semantic understanding is required in the change detection research such as disaster investigation. The paper proposes the concept of \textit {semantic change detection}, which involves intuitively inserting semantic meaning into detected change areas. We mainly focus on the novel semantic segmentation in addition to a conventional change detection approach. In order to solve this problem and obtain a high-level of performance, we propose an improvement to the hypercolumns representation, hereafter known as hypermaps, which effectively uses convolutional maps obtained from convolutional neural networks (CNNs). We also employ multi-scale feature representation captured by different image patches. We applied our method to the TSUNAMI Panoramic Change Detection dataset, and re-annotated the changed areas of the dataset via semantic classes. The results show that our multi-scale hypermaps provided outstanding performance on the re-annotated TSUNAMI dataset.
\end{abstract}


%
\IEEEpeerreviewmaketitle

\section{Introduction}

Change detection is the study of detecting changes between two different images of the same scene taken at different times. The main task is to distinguish significant differences between images from irrelevant background information such as illumination variations and viewpoint changes. The current focus of change detection is aimed at using the process to detect changes in images taken before and after natural disasters in order to facilitate the reconstruction of buildings in affected cities. In the computer vision field, convolutional neural networks (CNNs) are especially useful for facilitating image recognition tasks through the use of convolutional, max-pooling, and fully-connected layers~\cite{KrizhevskyNIPS2012}. Additionally, Sakurada \textit{et al.} applied deeper CNN architecture and considered region fitting based on superpixel segmentation~\cite{AchantaPAMI2012} and context geometry~\cite{HoiemICCV2005}. The approach in ~\cite{SakuradaBMVC2015} is robust to irrelevant differences such as those resulting from weather and ground variations.

We must consider the ``where and how" differences between two images taken at different times, rather than just the ``where" difference, which is the conventional change detection problem. On the other hand, since extracting semantic understanding of change areas is a time-consuming task for human beings, we thought to solve the \textit{semantic change detection (SCD)} problem by inserting semantic definitions into changed areas.

\begin{figure*}[t]
\begin{center}
   \includegraphics[width=1.0\linewidth]{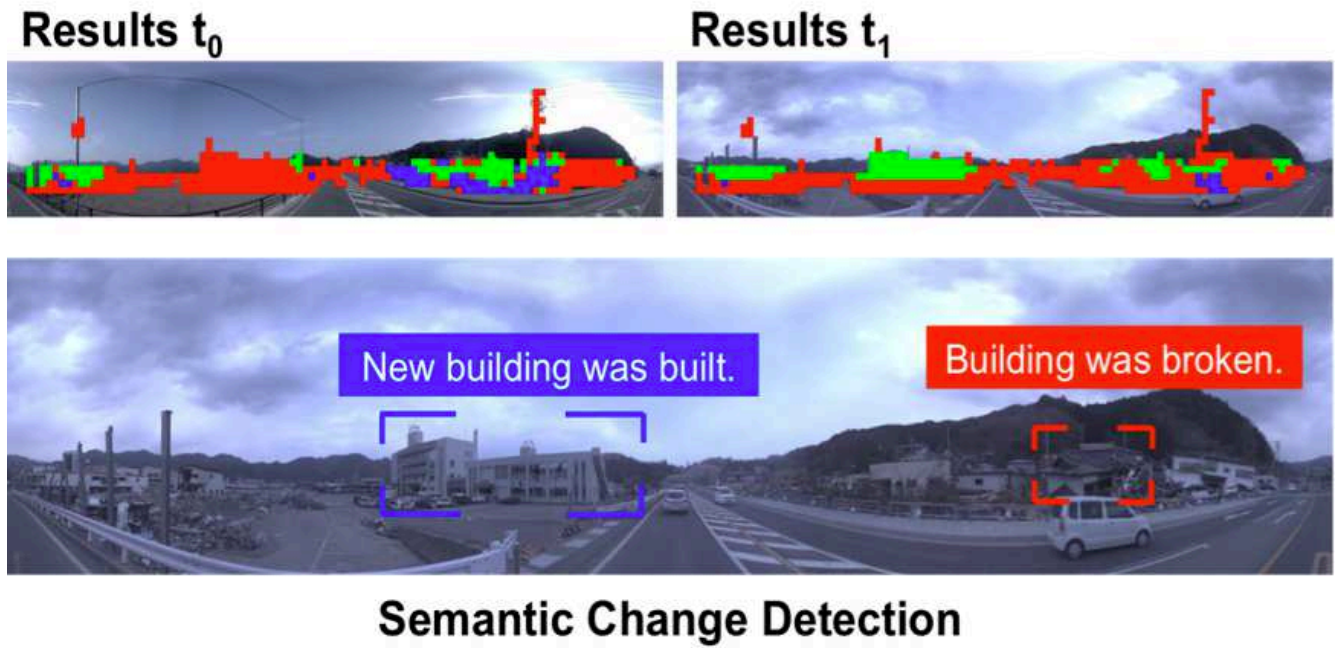}
\end{center}
   \caption{Concept of semantic change detection}
\label{fig:scd}
\end{figure*}

The two main contributions of this paper are as follows:
\begin{itemize}
  \item We propose a novel concept \textit{semantic change detection} that intuitively assigns semantic meaning to detected change areas by solving a two-part problem that consists of semantic segmentation and change detection.~\footnote{To achieve a novel concept for semantic change detection, we mainly    improve the semantic segmentation approach to update the concept. In the change detection, we employ a conventional method. } Figure~\ref{fig:scd} illustrates the concept which is to understand a changed area and its semantic meaning. We then apply our method to the TSUNAMI Panoramic Change Detection dataset~\cite{SakuradaBMVC2015}, which re-annotated the changed areas in the dataset as semantic classes.
  \item Hypercolumns~\cite{HariharanCVPR2015} are effectively incremented in order to solve the problem of semantic change detection. Originally, hypercolumns were used to represent pixel-wise low-level outputs of a feature map. However, in our method, we accumulate region-based values for each feature map. The results of our experiments show that our region-based approach is simple, yet effective.
\end{itemize}

The rest of this paper is organized as follows. In Section 2, related works are listed. The definitions of semantic change detections and solutions are shown in Sections 3 and 4, respectively. The experimental results and considerations are shown in Section 5. Finally, we conclude the paper in Section 6.

\section{Related works}
\subsection{Change detection}

Change detection is defined as capturing the differences in two different images of the same scene that were taken separately. The difficulties related to this process are normally the result of light source variations and viewpoint changes. Accordingly, irrelevant changes to the semantic context must be eliminated. To accomplish this, Gueguen \textit{et al.} proposed a method to detect damage from aerial images~\cite{GueguenCVPR2015} that involved the creation of a bag-of-words vector from assigned hierarchical local features. Furthermore, in their study, Sakurada \textit{et al.} achieved their intended change detection approach on the TSUNAMI dataset~\cite{SakuradaBMVC2015}, and the fine-tuned CNN model combined with the SUN database~\cite{XiaoCVPR2010} was found to strengthen the task of change detection when applied to natural scenes and artificial building regions. Moreover, they executed change detection with activated CNN features that combined superpixels~\cite{FelzenszwalbIJCV2004} and geometric contexts~\cite{HoiemICCV2005}. Furthermore, methods for extracting semantic meaning from the changed areas were presented in Sakurada's work~\cite{SakuradaBMVC2015}. 

The change detection framework must be improved with semantic meaning for a practical application. A user can easily understand a situation, where and how different each other between two images. (see Figure~\ref{fig:scd})

\subsection{Convolutional neural networks}

The original CNN explained in LeCun's LeNet-5 space displacement neural network (SDNN)~\cite{LeCunIEEE1998} was used for 10-digit character recognition. LeNet-5 is based on Neocognitron~\cite{FukushimaBC1980}. \cite{LeCunIEEE1998} contains convolutional, max-pooling and fully-connected layers, which are the basic types of CNN architecture. The famous deep model proposed by Alex Krizhevsky~\cite{KrizhevskyNIPS2012} is called AlexNet. Two years after the creation of AlexNet, deeper CNNs, such as GoogLeNet~\cite{SzegedyCVPR2015} and Oxford's Visual Geometry Group Network (VGGNet)~\cite{SimonyanICLR2015} were proposed. The GoogLeNet consists of inception units that have several convolutional parameters. In total, GoogLeNet has 22 layers with nine inception units. The VGGNet basically connects 16/19 layers convolution and fully-connected layers. The VGGNet performs max-pooling after a couple of convolutional layers with a static patch size of $3\times3$ in order to create sophisticated non-linearity in a network. In the 2015 ImageNet Large Scale Visual Recognition Challenge (ILSVRC2015), Microsoft proposed deep residual networks (ResNet)~\cite{HeCVPR2016resnet} for image recognition, object detection, and semantic segmentation. The total architecture of their proposal is 152 layers high, and its unique characteristic is its ability to stack residual learning at three layer intervals. 

In a different approach, Donahue \textit{et al.} demonstrated the effectiveness of transfer learning with activated CNN features and linear classifiers~\cite{DonahueICML2014} and showed that fully-connected layers could be effective for transferring data. However, even though \cite{DonahueICML2014} assigned the AlexNet architecture, the most recent works~\cite{SakuradaBMVC2015, YangICCV2015} employ VGGNet. Accordingly, we decided to apply the VGGNet net to our semantic learning method.

\subsection{Semantic segmentation}

The semantic segmentation task is more difficult than the image recognition and object detection tasks because the problem must deal with pixel-level multi-class categorization. Given a region-based small patch, the semantic segmentation approach returns a class level for each pixel.

TextonBoost, which was an early approach in semantic segmentation~\cite{ShottonIJCV2007}, produces comprehensive judgments by optimizing conditional random fields (CRF). Later, Mostajabi \textit{et al.} applied a multi-scale feature in addition to a superpixel segmentation~\cite{MostajabiCVPR2015}. However, more effective approaches are proposed in the FCN~\cite{LongCVPR2015}, SegNet~\cite{BadrinarayananarXiv2015} and hypercolumns~\cite{HariharanCVPR2015}. Especially in the hypercolumns give a jointly flexible and effective representation for the task of semantic segmentation. The concatenated vector (2nd pool, 4th conv, 7th fc layers) allows us to represent low-, mid-, and high-level features for different parts of the semantic segmentation. In this paper, we aim at an improved method of vector creation for semantic change detection based on hypercolumns and multi-scale feature representation.

\section{Semantic change detection}

\begin{figure}[t]
\begin{center}
   \includegraphics[width=1.0\linewidth]{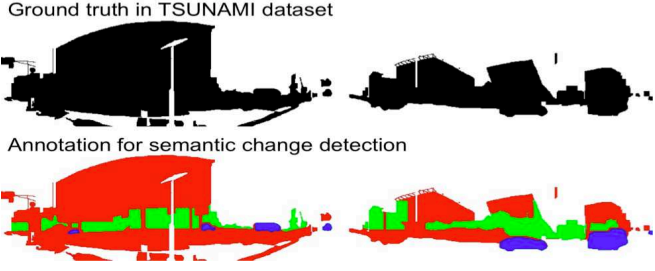}
\end{center}
   \caption{Original annotation in the TSUNAMI dataset~\cite{SakuradaBMVC2015} (top) and semantic change detection annotation (bottom): Three labels were inserted into the dataset: car (blue), building (green), and rubble (red)}
\label{fig:annotation}
\end{figure}

\textit{Semantic change detection} involves applying semantic meaning to intuitively detected change areas by solving a two-part problem consisting of semantic segmentation and change detection. Figure~\ref{fig:annotation} shows an example of semantic change detection annotation. In the TSUNAMI dataset, 200 images consisting of 100 pairs are utilized. Each pair consists of images taken before and after disasters at times $t0$ and $t1$, respectively. The three semantic labels are listed in the re-annotated dataset as $L_{i} = \{L_{0}, L_{1}, L_{2}\}$. 

We inserted three semantic labels into the changed areas in the TSUNAMI dataset, which we named car (blue), building (building), and rubble (red). It is obvious that the car and building classes are important in the aftermath of a disaster. Additionally, we set another class that consisted of rubble. By separating the building and rubble classes, we could quickly grasp situations in which a building disappeared after a disaster. However, some difficult points will remain in semantic change detection if only three classes are used. These difficulties are:
\begin{itemize}
  \item We must execute pixel-level semantic evaluation when given a small-patch.
  \item Sometimes a patch will have textureless regions.
  \item Sometime inter-class appearances are too close (e.g. building and building rubble).
\end{itemize}


\section{Hypermaps representation}

Hypercolumns effectively implement semantic segmentation by concatenating lower layers into high-level, fully-connected activated features~\cite{HariharanCVPR2015}. Here, we insert convolutional map information into the lower layers feature. Figure~\ref{fig:hypermaps} shows a comparison between hypercolumns and our hypermaps. Moreover, we implement hypermaps with multi-scale representation. The details are shown as follows:

\subsection{Hypercolumns}

In the first step, the hypercolumns access the feature maps. Although the fully-connected layer consists of flattened elements, the convolutional or pooling layers are listed as feature maps. Since feature maps are usually smaller than the size of the input image, Hariharan \textit{et al.} upsampled the feature maps in order to keep them the same size as the input image. The original hypercolumns concatenate pool2, conv4, and fc7 from AlexNet~\cite{KrizhevskyNIPS2012}. However, our more sophisticated feature map is based on VGGNet~\cite{SimonyanICLR2015}. Next, we upgraded to the high-level CNN architecture. The concatenated vectors are pool2 (128 channels), conv4\_3 (512 channels), and fc7 (4096 channels) which results in 4,736 dimensions. 

\subsection{Hypermaps (ours)}

The hypermaps concept is based on extracting a representative value per feature map, specifically 128 on pool2 and 512 on conv4\_3 (Figure \ref{fig:hypermaps} right). The resulting representation enables us to comprehensively understand the patch-based feature. We accumulate feature map values using the simple process described below:
\begin{eqnarray}
  f'_{k} &=& \sum_{i=1}^{W*H} f_{ik}
\end{eqnarray}
where $f'_{k}$ is a representative value at the feature map $f_{k}$, $W$ and $H$ represent the width and height of the feature map, respectively. 

Moreover, since the feature map center should be a high-weighted value, we generate the required weighted value based on the Gaussian distribution.
\begin{eqnarray}
  f'_{k} &=& \sum_{i=1}^{W*H} \alpha_{i} f_{ik} \\
  \alpha &\sim& N(\mu, \sigma^{2})
\end{eqnarray}
Here, $\mu$ is the feature map center.

Against to the previous hypercolumns~\cite{HariharanCVPR2015}, our hypermaps representation enables to extract a wider range feature map around the focused point. 

\subsection{Multi-scale representation}

\begin{figure}[t]
\begin{center}
   \includegraphics[width=0.80\linewidth]{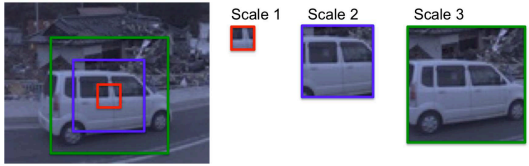}
\end{center}
   \caption{Multi-scale representation}
\label{fig:multiscale_patch}
\end{figure}

Multi-scale map extraction provides an improved representation for determining semantic meaning (see Figure~\ref{fig:multiscale_patch}). Here, we prepare three patches for the same regions to use in our evaluations. The three patches are different in size from each other. A label $L_{i}$ is assigned for each patch size, after which the maximum count $C_{i}$ is used as the semantic meaning.
\begin{eqnarray}
  y &=& max_{i \in [1,N]} C_{i}
\end{eqnarray}
$N$ is the number of labels. In our experimental section, we evaluate patch sizes of $\{10, 30, 50, 70, 90\}$ [pixel] . We then apply support vector machines (SVMs) as linear classifiers.

\subsection{Other representations}

In the experimental section, we will compare our method to other approaches.

Two representative models were employed from hand-crafted features, namely SIFT+BoW~\cite{CsurkaECCVW2004} and histograms of oriented gradients (HOG)~\cite{DalalCVPR2005}. Here, we cite the Deep Conventional Feature (DeCAF) paper~\cite{DonahueICML2014}, which explores transfer learning by using the CNN activation feature and the SVM classifier. This paper states that use of the first fully-connected layer provides the best way to accomplish transfer learning. Although \cite{DonahueICML2014} preferred AlexNet~\cite{KrizhevskyNIPS2012}, we replaced the network architecture with VGGNet~\cite{SimonyanICLR2015}. We also evaluated Microsoft's deep residual networks (ResNet)~\cite{HeCVPR2016resnet}, which were the ILSVRC2015 winner. The ResNet provides an optimized residual of convolutional maps in the training step. We applied the fifth pooling layer (pool5; 2048 [dimensions]) and the fully-connected layer (fc; 1000 [dimensions]) obtained from the ImageNet pre-trained model. 


\begin{figure}[t]
\begin{center}
   \includegraphics[width=0.80\linewidth]{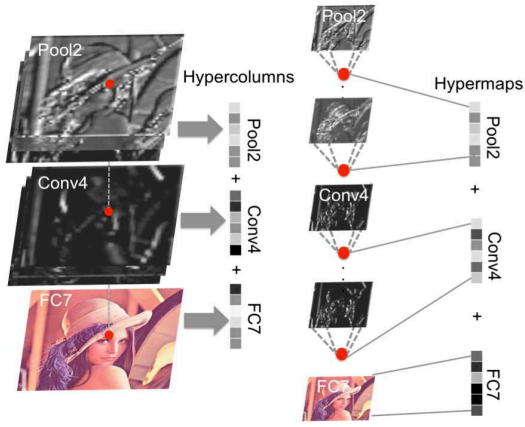}
\end{center}
   \caption{Hypercolumns and our hypermaps}
\label{fig:hypermaps}
\end{figure}

\section{Experiments}

\begin{figure}[t]
\centering
\subfigure[]{\includegraphics[width=1.0\linewidth]{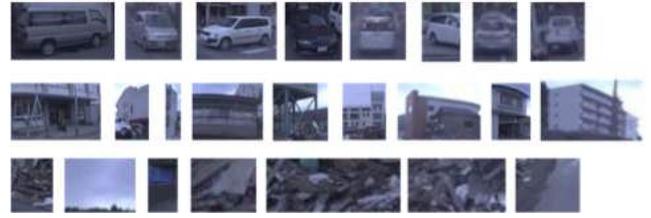} \label{fig:learningimage}}
\\
\subfigure[]{\includegraphics[width=1.0\linewidth]{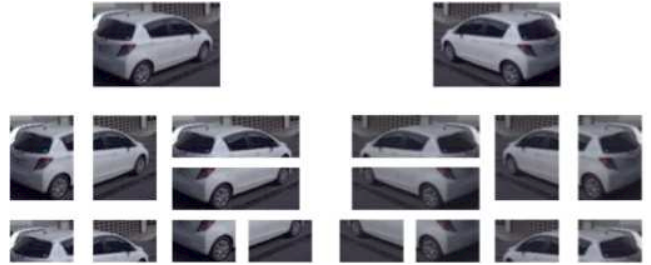}
\label{fig:dataaugx18}}
\\
\caption{Learning samples: (a) patch-based learning samples (b) data augmentation}
\label{fig:learningsample}
\end{figure}

\begin{figure}[t]
\centering
\subfigure[]{\includegraphics[width=0.45\linewidth]{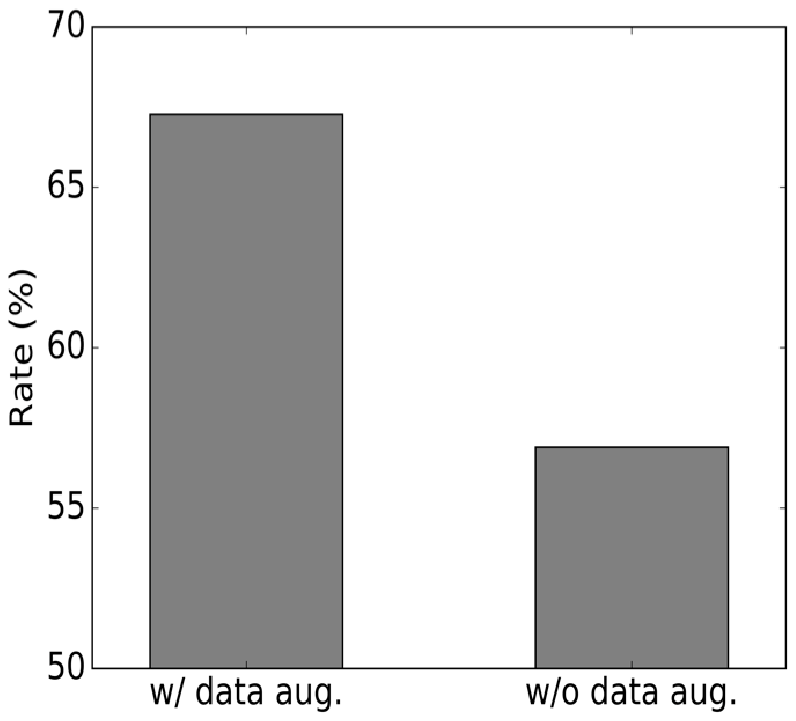} \label{fig:dataaug}}
\subfigure[]{\includegraphics[width=0.45\linewidth]{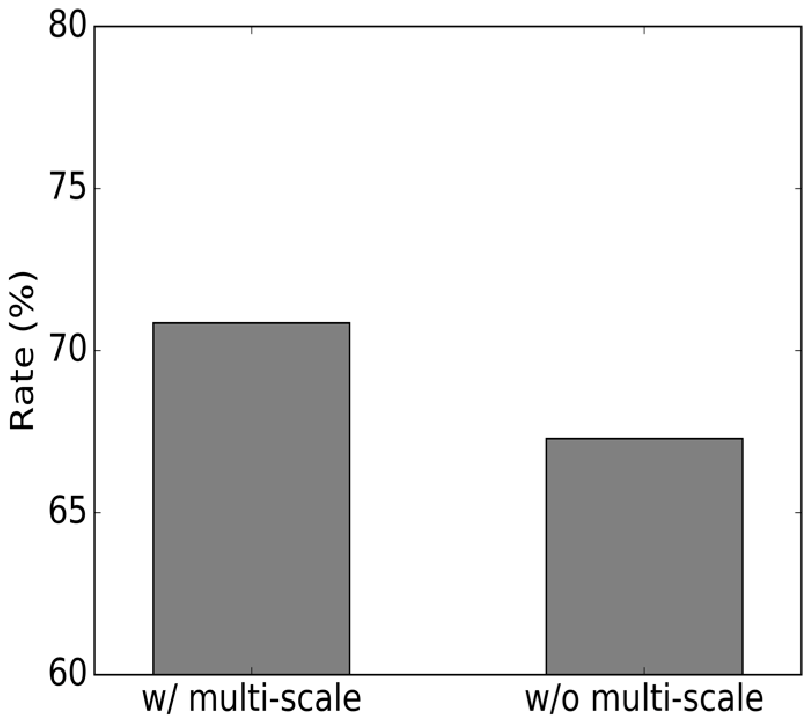}
\label{fig:multiscale}}
\caption{(a) With and without data augmentation (b) With and without multi-scale representation}
\label{fig:parameter_ms_data}
\end{figure}

\begin{figure}[t]
\begin{center}
   \includegraphics[width=0.80\linewidth]{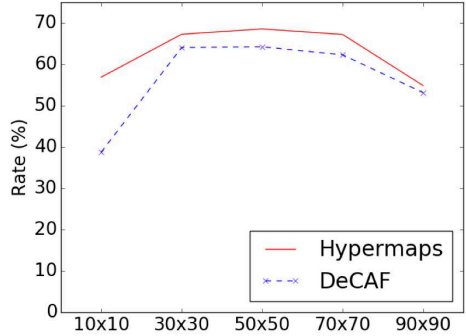}
\end{center}
   \caption{Relationship between patch size and accuracy}
\label{fig:patchsize}
\end{figure}

\subsection{Learning and testing}

Patches with the $L_{i} = \{L_{0}, L_{1}, L_{2}\}$ label were cropped. The number of learning data consisted of the following $t_{0}$: car 280, building 537, and rubble 701, $t_{1}$: car 352, building 631, and rubble 921. Although the learning examples are shown in Figure~\ref{fig:learningsample}(a), we executed data augmentation ($\times$18) using image flip and division (Figure~\ref{fig:learningsample}(b)). Learning and testing are performed for cross-validation. More specifically, $t0$ testing is provided from $t1$ patch learning and $t1$ testing is provided from $t0$ patch learning. The testing offset is set as $(x,y)=(10,10)$ [pixel].

\subsection{Parameter tuning}

A parameter tuning evaluation was conducted, the elements of which are listed below:

\begin{itemize}
  \item w/ or w/o data augmentation (Figure~\ref{fig:dataaug}; w/ data aug. is better).
\end{itemize}

Accuracy levels, with and without data augmentation, are shown in Figure~\ref{fig:dataaug}. The effect of data augmentation is clear from the results of with (67.28\%) or without (56.90\%). Accordingly, we will employ x18 data augmentation hereafter.

\begin{itemize}
  \item Multi-scale feature patch size (Figure~\ref{fig:patchsize}; $30\times30$, $50\times50$ and $70\times70$ should be combined).
\end{itemize}

Figure~\ref{fig:patchsize} shows the relationship between patch size and accuracy. We adjusted the parameter from $10\times10$ to $90\times90$ at each 20-pixel increment. According to the figure, the multi-scale feature should be fixed at patch sizes of $30\times30$, $50\times50$, and $70\times70$ [pixel] (which provides better performance than using all five patches). These three patch types allow sufficiently higher levels of performance and provide the same level of semantic change detection accuracy. Therefore, we set the basic patch size as $30\times30$ [pixel] for conventional approaches.

\begin{itemize}
  \item w/ or w/o multi-scale representation (Figure~\ref{fig:multiscale}; w/ multi-scale representation is better).
\end{itemize}

With and without multi-scale representation is shown in Figure~\ref{fig:multiscale}. Here, it can be seen that the multi-scale representation thoroughly evaluates the center of an image patch. Especially in the situations where the panoramic change detection dataset is used, it is clear that multi-scale representation should be implemented due to the changing scale of the dataset. The rate change was measured at 67.28\% (without multi-scale) to 70.85\% (with multi-scale).

\begin{figure}[t]
\begin{center}
   \includegraphics[width=0.85\linewidth]{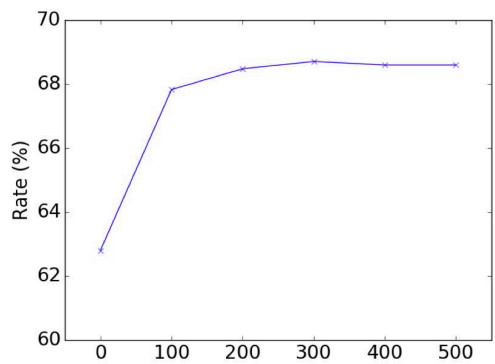}
\end{center}
   \caption{Gaussian parameter $\sigma^{2}$}
\label{fig:sigma}
\end{figure}

\begin{itemize}
  \item Gaussian parameter $\sigma^{2}$ (Figure~\ref{fig:sigma}; $\sigma^{2}=300$ is the best).
\end{itemize}

Next, tuning was performed to fix the Gaussian parameter $\sigma^{2}$ in the hypermaps. Here, we assigned a single patch size ($70\times70$). The best rate was obtained when $\sigma^{2}=300$.

\begin{itemize}
  \item w/ or w/o weighted value for hypermaps (Table~\ref{tab:weightedvalue}; we apply w/ weighted value).
\end{itemize}

The last parameter to be examined was with or without the weighted value for hypermaps, and the results of that evaluation are shown in Table~\ref{tab:weightedvalue}. Although the performance rates are very close, it can be seen that the with weighted value is better in all cases than the without weighted value for the dataset utilized here.

\begin{figure}[t]
\centering
\subfigure[]{\includegraphics[width=0.48\linewidth]{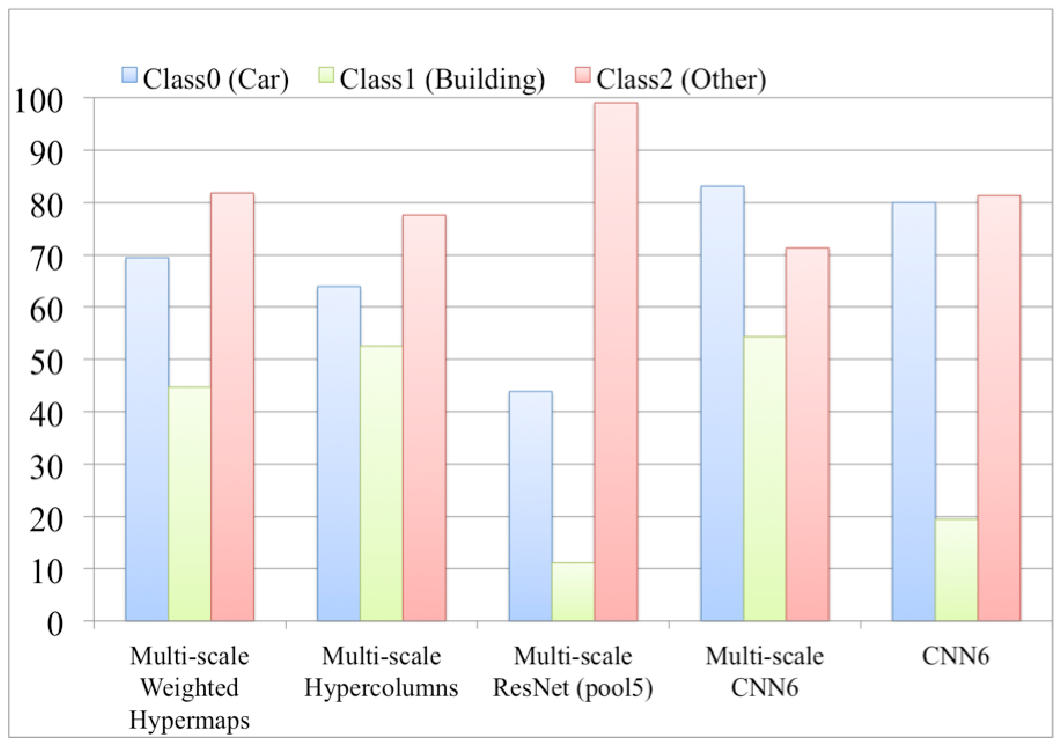} \label{fig:eachclasst0}}
\subfigure[]{\includegraphics[width=0.48\linewidth]{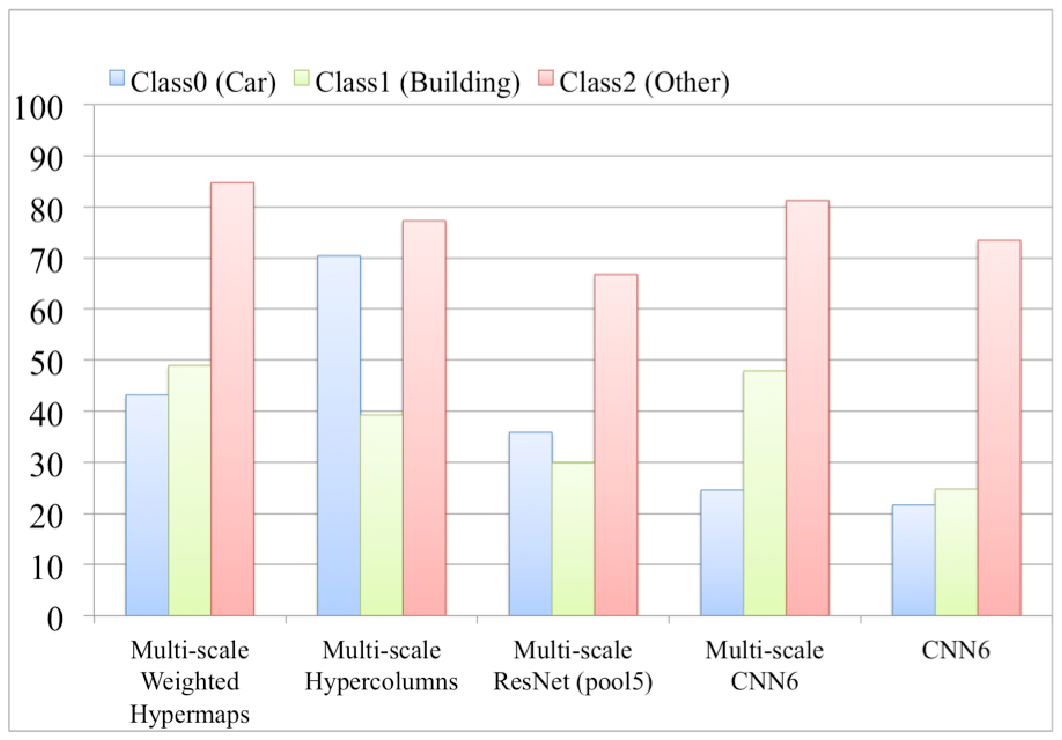}
\label{fig:eachclasst1}}
\\
\caption{Performance evaluation for each class on the (a) $t_{0}$ and (b) $t_{1}$ sets: multi-scale w/ and w/o weighted hypermaps (ours), multi-scale hypercolumns, multi-scale ResNet (pool5), multi-scale CNN6, CNN6}
\label{fig:eachclass}
\end{figure}

\begin{table}[t]
\begin{center}
\caption{With or without the weighted value for hypermaps}
\begin{tabular}{|l|c|c|}
\hline
 & \% on $t_{0}$ & \% on $t_{1}$ \\
\hline
w/ weighted value (ours) & 71.18 & \textbf{66.44} \\
\hline
w/o weighted value & \textbf{71.19} & 65.93 \\
\hline
\end{tabular}
\label{tab:weightedvalue}
\end{center}
\end{table}

\begin{table}[t]
\begin{center}
\caption{Results for the re-annotated TSUNAMI dataset}
\begin{tabular}{lcc}
Approach & $t_{0}$ (\%) & $t_{1}$ (\%) \\
\hline
SIFT+BoW~\cite{CsurkaECCVW2004} & 53.60 & 39.40 \\
HOG~\cite{DalalCVPR2005} & 63.70 & 52.30 \\
DeCAF~\cite{DonahueICML2014} & 64.08 & 50.61 \\
Multi-scale DeCAF~\cite{DonahueICML2014} & 66.88 & 64.60 \\
Multi-scale ResNet (fc)~\cite{HeCVPR2016resnet} & 63.10 & 40.10 \\
Multi-scale ResNet (pool5)~\cite{HeCVPR2016resnet} & 69.68 & 48.17 \\
Multi-scale Hypercolumns~\cite{HariharanCVPR2015} & 66.54 & 62.90 \\
Multi-scale Weighted Hypermaps (ours) & \textbf{71.18} & \textbf{66.44} \\ 
\hline
\end{tabular}
\label{tab:comparison}
\end{center}
\end{table}

\subsection{Results for the re-annotated TSUNAMI dataset}

A comparison of the re-annotated TSUNAMI dataset results is shown in Table~\ref{tab:comparison}. Here, it can be seen that our multi-scale hypermaps achieved the best performance rate in the re-annotated TSUNAMI dataset. The detailed consideration is listed as follows:

The CNN feature, DeCAF~\cite{DonahueICML2014} is better than the hand-crafted feature SIFT+BoW~\cite{CsurkaECCVW2004}. CNN enables us to describe general image features due to the use of the ImageNet pre-trained model, which contains more than 1.0 M training images that have wide-ranging variations. The data-driven parameter tuning allows us to derive a sophisticated feature for image recognition. The transfer learning method based on DeCAF is effective for assigning semantic meaning to changed areas. The HOG feature~\cite{DalalCVPR2005} performs at nearly the same level as DeCAF, but the $30\times30$ patch size is more suitable to the problem under discussion. Finally, unlike the HOG feature, we noted that the CNN feature has a location-free property due to the fully-connected layer. 

Next, we implemented multi-scale representation with a couple of patch sizes. The effectiveness is shown in Table~\ref{tab:comparison} where the differences between DeCAF and multi-scale DeCAF, with and without multi-scale representation, can be clearly seen. The performance rates improved +2.38\% on the $t_{0}$ and +13.83\% on the $t_{1}$. The data at time $t_{1}$ was more difficult since rubble was included in $L_{2}$ (other class). Since the system was required in order to evaluate $t_{0}$ training, which does not contain rubble in class $L_{2}$, it is clear that the multi-scale representation is effective for semantic change detection. 

The ResNet achieved an outstanding rate during the ILSVRC2015, but the network architecture does not fit the semantic change detection problem, primarily because it is unsuitable for transfer learning from the fully-connected layer and average-pooling layer. The activated features are focused on the 1,000 categories of the ImageNet database. The results of multi-scale ResNet were 63.10\% ($t_{0}$) and 40.10\% ($t_{1}$) with the fully-connected layer, and 69.68\% ($t_{0}$) and 48.17\% ($t_{1}$) with the fifth max pooling layer.

The difference between hypercolumns and hypermaps (ours) is shown in Figure~\ref{fig:hypermaps}. As can be seen in the figure, the hypermaps accumulate surrounding values in order to improve the semantic change detection task result. Here, we handle the weighted function with the Gaussian distribution. Since the performance rate was +4.65\% ($t_{0}$) and +3.03\% ($t_{1}$) over the hypercolumn representation values, it is clear that our proposal produced the best percentage for the $t_{0}$ and $t_{1}$ data, and that both multi-scale and hyper map representation were remarkably capable in terms of assigning semantic meaning to changed areas.

\begin{figure}[t]
\begin{center}
   \includegraphics[width=1.0\linewidth]{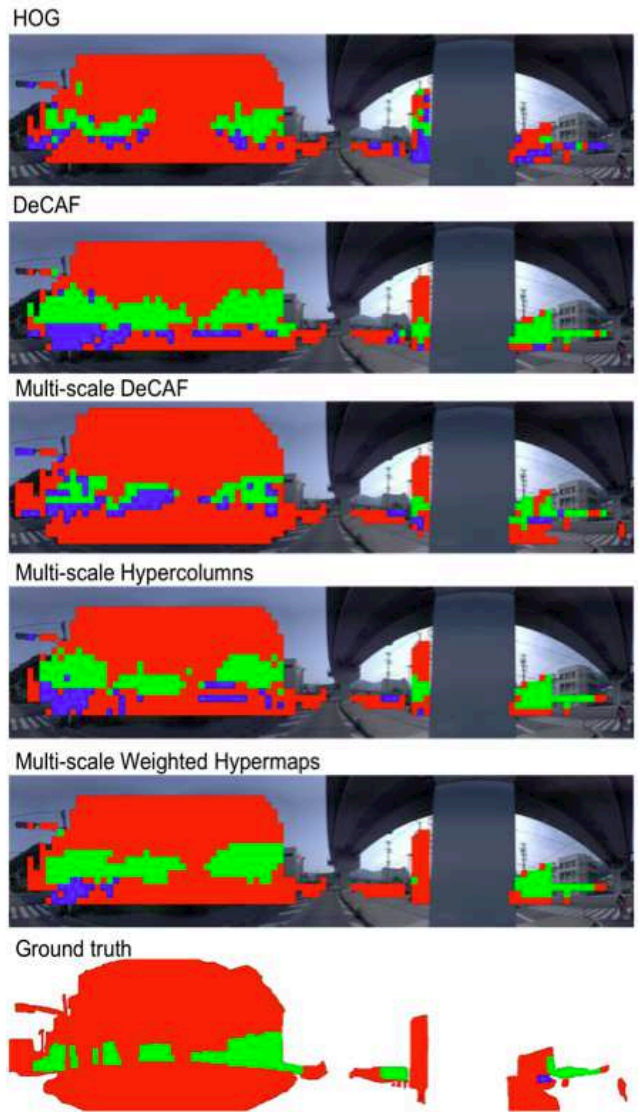}
\end{center}
   \caption{Re-annotated TSUNAMI dataset comparison results: from top to bottom, HOG, DeCAF, Multi-scale DeCAF, Multi-scale hypercolumns, Multi-scale weighted hypermaps, and ground truth}
\label{fig:compare}
\end{figure}

Figure~\ref{fig:eachclass} shows the performance rate on the $t_{0}$ (Figure~\ref{fig:eachclasst0}) and  $t_{1}$ (Figure~\ref{fig:eachclasst1}). Our multi-scale weighted hypermaps provide a balanced representation for all classes. Figure~\ref{fig:compare} shows the examples of semantic segmentation into the changed areas. The multi-scale weighted hypermaps (ours) gives a sophisticated representation by comparing with other segmentation algorithms.


\section{Conclusion}

In this paper, we proposed the concept of \textit{semantic change detection}, which recognizes the semantic meaning of changed areas. From an examination of conventional vision-based tasks, we determined that the primary difficulty related to change detection consists of the two-part problem of semantic segmentation and change detection. Our semantic change detection method allowed us to understand the ``where and how" differences between two images taken at time $t_{0}$ and $t_{1}$. 

The paper also shows that multi-scale hypermaps preformed remarkably well on the re-annotated panoramic change detection (TSUNAMI) dataset~\cite{SakuradaBMVC2015}. More specifically, the multi-scale hypermap records show 71.18\% on the $t_{0}$ and 66.44\% on the $t_{1}$. These rates are +4.64\% ($t_{0}$) and +3.54\% ($t_{1}$) above those obtained with the hypercolumns representation~\cite{HariharanCVPR2015}. 

In the future, we will attempt to implement end-to-end training with CNN, including (multi-) scale settings and feature representations. We also intend to extend data variation and data augmentation. It is also noted that longer observation periods will be required in order to properly evaluate this semantic change detection method.

\bibliographystyle{IEEEtran}
\bibliography{itsc17}

\end{document}